\pdfoutput=1
\documentclass[11pt]{article}

\usepackage{authblk}

\usepackage{acl}
\usepackage[ruled,vlined]{algorithm2e}
\usepackage{times}
\usepackage{latexsym}
\usepackage{graphicx}
\usepackage[T1]{fontenc}
\usepackage{subcaption}
\usepackage{booktabs}
\usepackage{amsmath}
\usepackage{url}
\usepackage[utf8]{inputenc}

\usepackage{microtype}

\usepackage{inconsolata}

\usepackage{amsfonts}
\usepackage{etoolbox}
\title{How Does the Textual Information Affect\\ the Retrieval of Multimodal In-Context Learning?}

\author[]{\textbf{Yang Luo}}
\author[]{\textbf{Zangwei Zheng}}
\author[]{\textbf{Zirui Zhu}}
\author[]{\textbf{Yang You}\vspace{-2ex}}
\affil[]{School of Computing, National University of Singapore\vspace{-2ex}}
\affil[]{\texttt{\{yangluo,zangwei,zirui,youy\}@comp.nus.edu.sg}}

\begin{document}
\maketitle
\begin{abstract}
% The increasing parameter size of multimodal large language models (MLLMs) is often accompanied by the emergence of various remarkable capabilities, and in-context learning is an important part of it, where MLLM achieves an increased downstream task ability without any update of pre-trained parameters. However, the performance of in-context learning is hugely affected by the selection of demonstrations. Existing methods only rely on images for the retrieval of context demonstrations but ignore the role of textual information. Moreover, there are several works discussing training a retriever for better retrieval of demonstrations for large language models, but the supervised retriever for MLLM remains unexplored.
% To solve this problem, we for the first time provide a comprehensive investigation of the impact of various modalities during the selection of in-context examples in the multimodal field and find that the retriever's performance is highly sensitive to the usage of different modalities. On the basis of the analysis, we further proposed a supervised MLLM-retriever that trains a neural network to choose examples
% that directly maximize multimodal in-context learning performance.  Extensive experiments on three different tasks demonstrate the efficiency of our method. 
% Besides, we also discussed the role of modalities in our training of supervised retrieval method and probed factors for the success of our model for further improvement in future work.

The increase in parameter size of multimodal large language models (MLLMs) introduces significant capabilities, particularly multimodal in-context learning, where MLLMs enhance task performance without updating pre-trained parameters. However, this effectiveness hinges on the appropriate selection of in-context examples, a process currently biased towards visual data, overlooking textual information. More importantly, the area of supervised retrievers for retrieval of multimodal in-context learning, crucial for optimal in-context example selection, continues to be uninvestigated.
Our study provides an in-depth evaluation of the impact of textual information on the unsupervised selection of in-context examples in multimodal contexts, uncovering a notable sensitivity of retriever performance to the employed modalities. Based on the above finding, we introduce a novel supervised MLLM prompt retriever \textbf{MSIER} that leverages a trained retriever based on MLLM's confidence to select examples, which enhances multimodal in-context learning efficiency. This approach is validated through extensive testing across three different tasks, demonstrating the method's effectiveness.
Additionally, we investigate the influence of modalities on our supervised retrieval method's training and explore the transferability of the supervised prompt retriever. This exploration paves the way for future advancements, highlighting the potential for refined in-context learning in MLLMs through the strategic use of multimodal data. The public code is available at \url{https://github.com/NUS-HPC-AI-Lab/Multimodal-ICL-Retriever}.
\end{abstract}

\section{Introduction}
\label{introduction}

The capability of large language models (LLMs) for in-context learning (ICL) has acquired significant attention and exhibited remarkable efficacy across a diverse range of downstream tasks \cite{an2023incontext, min2022rethinking, wang2023label}. ICL aims to adapt the pre-trained model to achieve high performance on downstream tasks without any update of parameters that require extra computing resources for the further training process. To be specific, 
the primary advantage of in-context learning lies in its capacity to assimilate and adapt to new information from its context. This is achieved by utilizing a series of question-and-answer pairs from the training dataset, known as in-context examples, which enable learning without necessitating updates to the model's parameters. Significantly, recent developments in multimodal large language models (MLLMs), which are constructed based on the foundations of large language models, have similarly exhibited capabilities in in-context learning \cite{alayrac2022flamingo, peng2023kosmos2, zhao2023mmicl,yin2023survey}, termed as multimodal in-context learning (M-ICL).
% Recent multimodal large language models (MLLMs), which are built upon LLMs, have also displayed such ICL ability \cite{alayrac2022flamingo, peng2023kosmos2, zhao2023mmicl}, termed as Multimodal In-Context Learning (M-ICL), as illustrated in Figure \ref{fig:micl}.

\begin{figure}
    \centering
    \includegraphics[width=0.38\textwidth]{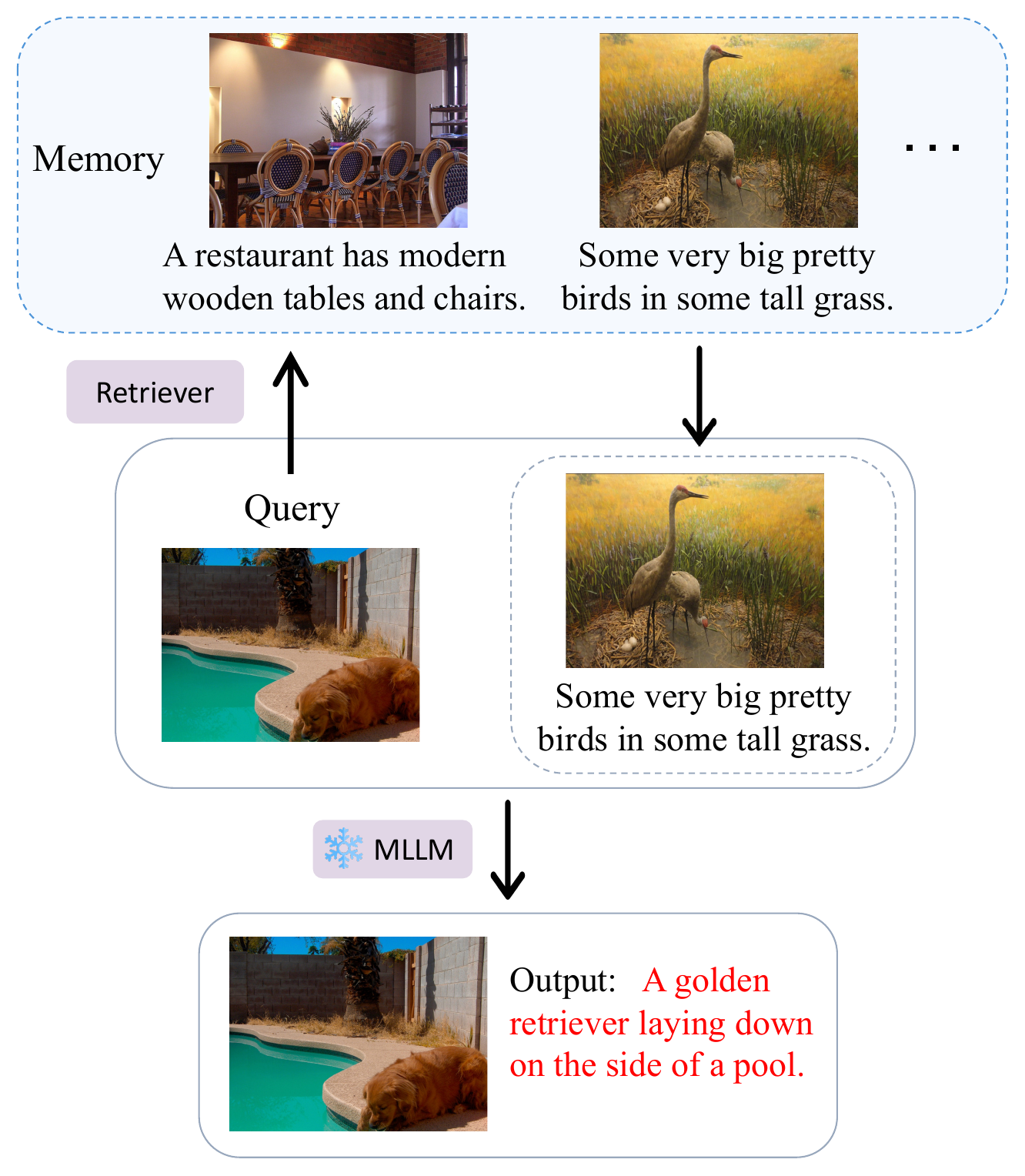}
    \caption{
% An overview of multimodal in-context example retrieval: upon receiving a query in the form of an image (image-text pair), which may comprise an image or a combination of an image and a textual question, from the test dataset, a retrieval mechanism identifies analogous examples within an indexed training dataset. 
% Subsequently, both the query and the identified training examples (collectively referred to as the prompt) are submitted to a multimodal large language model (MLLM) for the generation of the output.
An overview of multimodal in-context example retrieval: This process involves receiving an image or an image-text query from the test dataset, and then using a retrieval mechanism to find similar examples in a training dataset. These examples and the original query (collectively called the prompt) are then inputted into a MLLM to generate the output.
}
% }
    \label{fig:micl}
\end{figure}

A notable advantage of M-ICL lies in its ability to utilize a singular model across a variety of language comprehension tasks. Nevertheless, \cite{chen2023understanding, wu2023selfadaptive, ye2023compositional} revealed that the efficacy of this approach in downstream applications can differ significantly, influenced by the selection of in-context examples. This revelation has fueled interest in the concept of in-context example retrieval, as illustrated in Figure \ref{fig:micl}. In this process, training examples (\textit{memory}) for a given test instance (\textit{query}) are selected for the prompt, guided by a certain similarity metric. 
% For the retrieval of in-context examples for in-context learning of large language models, \cite{rubin2022learning} initially employed an unsupervised retrieval technique to select candidate examples, from which the top instances were identified using a supervised prompt retriever to enhance downstream performance.
% Contemporary studies \cite{yang2022empirical} in multimodality field have predominantly employed pre-existing unsupervised image similarity metrics or have developed specialized in-context example retrievers that choose examples based on their superficial resemblance.

In contrast to Natural Language Processing (NLP) \cite{dong2023survey, codaforno2023metaincontext}, where the primary input is text, multimodal tasks predominantly utilize a single image (image-text pair) as input and the memory for retrieval consists of multimodal examples. 
% Consequently, the primary challenge in multimodal in-context learning involves enabling the model to comprehend the prompt accurately and to make predictions based on a new image-text pair, referred to as the 'query pair'. 
% In pursuit of advancing this field, recent research () has introduced comprehensive frameworks for visual in-context learning \cite{zhang2023makes}. These frameworks involve selecting a relevant prompt from a database based on the query image, merging it with said query image to produce a composite image, and then inputting this amalgamated image into a large-scale model to generate a prediction. Other studies [31, 24, 23] have established that the quality of prompts significantly influences the effectiveness of language in-context learning. Within the domain of visual in-context learning, this raises a pertinent question: what are the essential factors that affect the performance of downstream tasks?
% Existing unsupervised retrieval of in-context examples of MLLM only considers image information but neglects the role of textual information, which raises a question about the effect of textual information in retrieval methods and whether more modalities will 
% improve the performance of in-context learning of MLLMs to some extent. Furthermore, when we are eager to construct a supervised retrieval framework, is it a critical factor to take the textual information seriously? Nevertheless, these problems receive little attention in multimodal in-context learning.
Current methodologies for unsupervised retrieval of in-context examples in multimodal large language models (MLLMs) predominantly focus on visual information\cite{alayrac2022flamingo, yang2022empirical, gu2023robust}, thereby overlooking the significance of linguistic data. This oversight prompts inquiries regarding the impact of textual content on retrieval processes and whether the inclusion of additional modalities could potentially enhance the efficacy of MLLMs' in-context learning capabilities. Moreover, 
% in the pursuit of developing a supervised retrieval framework, the question arises as to whether the integration of textual information constitutes a crucial element. 
the supervised retrieval framework for M-ICL has received minimal attention so far.

In this work, our attention is centered on M-ICL, an emerging concept in the MLLMs field with limited existing research on its practical application. Specifically, we undertook a thorough investigation into the effects of textual information for retrieval of in-context examples in M-ICL. Through this research, 
% we have identified a pivotal issue: the performance of the selection of in-context examples on downstream tasks is notably influenced by the choice of modality for both unsupervised and supervised retrievers. 
we have developed a supervised retriever that aligns with existing knowledge from MLLMs. This framework achieves high efficiency and demonstrates excellent transferability across various datasets and sizes of MLLMs.
The contributions of this paper are summarized as follows:

\begin{itemize}
    \item We present the first thorough analysis into the role of textual modality in both unsupervised and supervised retrieval of in-context examples for M-ICL and demonstrate that the addition of textual modality plays a crucial role in improving M-ICL performance.
    \item By considering both visual and textual information when selecting in-context examples, we design a Multimodal Supervised In-context Examples Retrieval (MSIER) framework with more efficient example selection performance by using a foundation MLLM scorer to evaluate the relevance and suitability of potential examples from multimodal data.
    % and explore the impact of text modal in the training of the proposed supervised retriever.
    \item Extensive experiments on three typical multimodal tasks demonstrate the high efficiency of our constructed supervised retrieval method that achieves the best performance. Besides, we further provide valuable insights regarding the importance and transferability of the supervised method in selecting highly relevant in-context examples for M-ICL.
\end{itemize}

\section{Related Work}
\label{related}
% Building upon the setting in (§3.1.2), at inference time, M-ICL can be implemented by adding a demonstration set, i.e., a set of in-context samples, to the original sample.

% \textbf{Retrieval} Recent scholarly interest in open-domain question answering has significantly accelerated research into dense retrievers, as evidenced by studies from Chen et al. (2017), Lee et al. (2019), Karpukhin et al. (2020), Guu et al. (2020), Khattab and Zaharia (2020), and Qu et al. (2021). Furthermore, the application of retrieval-based methods has expanded to encompass a broader range of knowledge-intensive tasks, such as fact verification, as explored by Lewis et al. (2020) and Samarinas et al. (2021). In a related vein, Pasupat et al. (2021) have also proposed the utilization of retrieval in semantic parsing, albeit with a specific focus on controlling the output generated by the model. Additionally, retrieval methods have demonstrated efficacy in the field of language modeling.

\textbf{In-context Examples Retrieval in NLP and CV} 
The field of natural language processing has identified that the choice of in-context examples significantly influences performance, as evidenced by \cite{agrawal2022incontext} and \cite{min2022rethinking}. 
Furthermore, the construction of these in-context examples, often referred to as prompts, including aspects such as relevance and diversity of retrieved examples, has been reported to impact performance as well. This understanding has guided the community toward investigating optimal methods for selecting effective in-context examples for large language models. In their study, \cite{liu2021makes} posited that effective in-context examples should bear semantic similarity to query sentences. Based on this hypothesis, they advocated for the selection of the nearest neighbors in the training set, as determined by a sentence encoder like RoBERTa \cite{liu2019roberta}. \cite{rubin2022learning} initially employed an unsupervised approach to gathering potential candidates, from which the most suitable examples were selected using a supervised method. 

Correspondingly, \cite{zhang2023VisualPromptRetrieval} explored the application of a supervised retriever within the scope of visual in-context learning, demonstrating enhanced performance outcomes. Nevertheless, the significance of textual information is overlooked within unsupervised and supervised retrievals for M-ICL, and the impact of various modalities on the development of the supervised retriever continues to be an area requiring investigation.

\textbf{Multimodal Retrieval-augmented Generation}
Retrieval augmentation for multimodal models enhances their capabilities by infusing externally retrieved information into their workflow. \cite{chen2022murag} proposed MuRAG, which accesses an external multimodal memory to augment language generation. \cite{yasunaga2023retrievalaugmented} presented the first retrieval-augmented multimodal model that enables a base model to refer to relevant text and images fetched from external memory. \cite{ramos-etal-2023-retrieval} introduced a new approach to image captioning that generates sentences given the input image and retrieved captions. 
% However, these methods require further training or fine-tuning to synthesize aspects from both the input and retrieved information. On the contrary, the in-context examples provided for M-ICL are used as part of the input sequence to the model, without deeper integration into the model's architecture that requires further parameter updates, allowing the model to interpret and generate based on its existing capabilities.
The retrieval process in RAG is designed to retrieve supplementary information to enhance the context for text generation and to produce more informed and contextually rich outputs. On the other hand, the retrieval of in-context examples in M-ICL aims to find examples that demonstrate how to perform a given task, providing explicit guidance to the model on how to generate its output. 
However, the role of textual information in the task of in-context example retrieval has remained unexplored so far.
% Despite the importance of optimal in-context example selection for MLLMs, the area of using supervised retrieval methods tailored for this task has remained largely unexplored so far.

\textbf{In-context Learning} In-context learning was first proposed by \cite{brown2020language} in their paper introducing GPT-3. It is a significant departure from the traditional learning method based on stochastic gradient descent and does not involve any parameter updates: the model needs to be provided with a few examples of the task as part of the context for text generation.
The size of the model and training data were thought to be key to training a model with in-context learning capabilities. More recently, there has been more research on the exact causes of in-context learning. \cite{min2022metaicl} has proposed MetaICL, a meta-training framework to elicit in-context learning capabilities in text-only language models. MetaICL conditions each example with related in-context examples during training. 
% Chan et al. [5] have investigated the distributional properties of training data for in-context learning. 
% Their findings have shown that certain properties encourage in-context learning in language models, and massive textual data from the web used to train large language models naturally have those properties.
In the context of MLLMs, ICL has been extended to more modalities, leading to Multimodal ICL\cite{alayrac2022flamingo, li2023otter, chen2023sinc}. 

% More related work on in-context learning can be found in Appendix \ref{sec:appendix::rw}.

\section{Method}
\label{method}

\begin{figure*}
    \centering
    \includegraphics[width=0.95\textwidth]{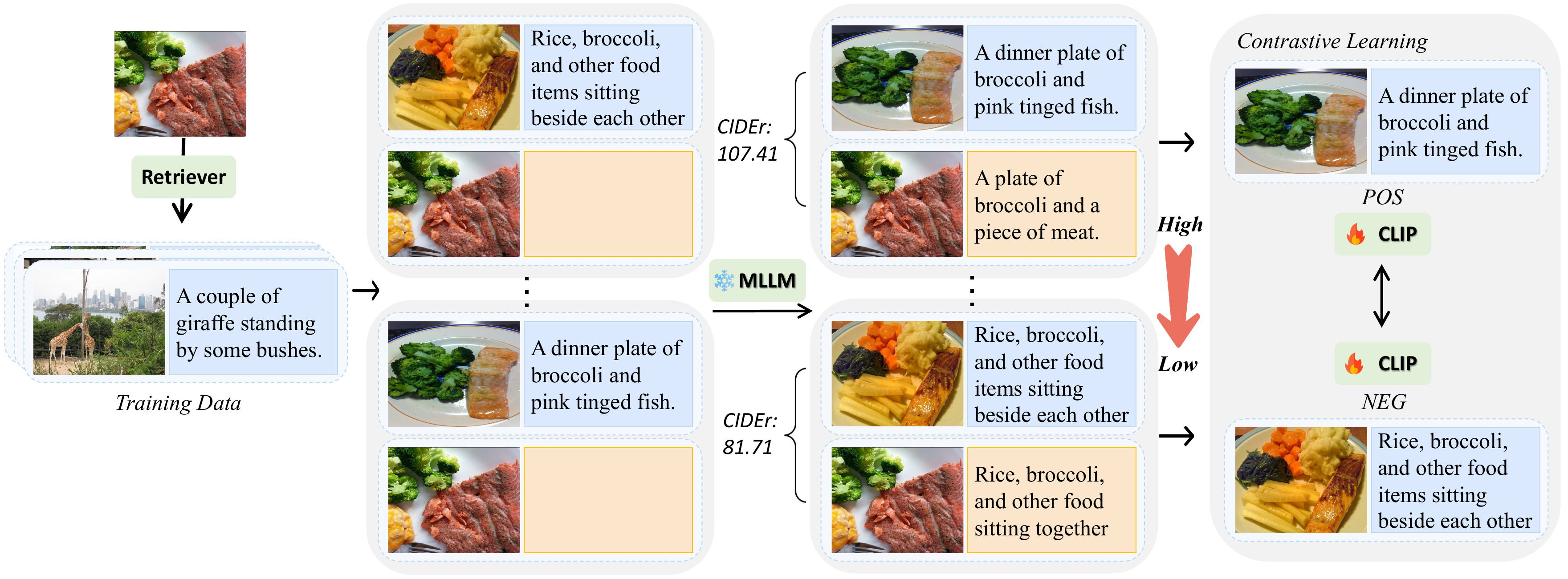}
    \caption{
Overview of the MSIER Method: The fundamental principle involves assessing the in-context learning performance for each source instance, thereafter identifying and choosing those instances exhibiting the most favorable or least favorable outcomes. These selected instances are then utilized to form a dataset, categorized as either positive or negative, which is essential for the facilitation of contrastive learning. The examples with high CIDEr scores (corresponding to \textbf{low NLL loss} during the scoring process) are selected as positive samples.}
    \label{fig:msier}
\end{figure*}

\subsection{Multimodal In-Context Learning}
% Emerging from the advent of large autoregressive language models pre-trained on extensive datasets, in-context learning represents a paradigm shift that obviates the need for parameter updates, distinguishing it markedly from conventional fine-tuning approaches. This methodology relies on generating conditional predictions through a set of exemplar input-output pairs, rather than adjusting the model's internal parameters. 
% Specifically, w
Within the realm of multimodal tasks such as image captioning, a pre-trained MLLM utilizes an image and its corresponding caption as an in-context example to delineate requirements of the task — essentially instructing the model on the expected form of input and output. Subsequently, when presented with a new image, the model can produce a more precise caption based on these learned patterns. Crucially, M-ICL maintains the parameters of the pre-trained model unchanged, thereby offering a more resource-efficient approach for tackling downstream tasks.
% Similarly, in the computer vision field, 

Regularly, given a memory (training dataset) $D={[x_i, y_i]}_{i=1}^N$ comprising $N$ pairs of multimodalities information and their corresponding labels, alongside a query example $x_q$ from the test dataset and the pre-trained model $f$, the process of in-context learning can be described as follows:
\begin{equation}
    y_q = f([C_p, x_q])
\end{equation}
where the context prompt $C_p$ consists of a sequence of input-output pairs from $D$, e.g., the input is an image and the output is its caption for the captioning task. Specifically, prompt $C_p$ serves as a contextual guide, directing the model to generate the optimal $y_q$ corresponding to query $x_q$, while avoiding any alterations to the parameters of the larger model.

\subsection{Importance of text information for unsupervised MLLM retrieval}
\label{impo}

As for the framework of the Multimodal Unsupervised In-context Examples Retrieval (MUIER) which is a prevalent retrieval methodology, our analysis initially focuses on the contribution of various modalities. Traditionally, existing MLLMs that present outstanding M-ICL capability have primarily utilized image data to facilitate M-ICL tasks. For instance, Flamingo \cite{alayrac2022flamingo} employed the Retrieval-based In-Context Example Selection (RICES) strategy \cite{yang2022empirical} for the identification of appropriate in-context examples, predominantly assessing the similarity between the query image and the images stored in the memory (training dataset). Nonetheless, this approach overlooks the significant role of textual information, which indeed influences the efficacy of M-ICL.

To validate the effectiveness of text-augmented MUIER, we performed a comprehensive comparison across diverse configurations of unsupervised retrievers, encompassing three main settings specifically designed for the image captioning task: 
    (1) \textbf{Q-I-M-I} (\textbf{Q}uery-\textbf{I}mage-\textbf{M}emory-\textbf{I}mage) indicates the case where only image information of image-text pairs preserved in the memory was applied for the retrieval of context. As described in RICES, we only consider the cosine similarity between the query image and the memory image based on vision features extracted by the unsupervised retriever.
    % \item \textbf{image and meaningless text} means the image in the memory is paired with a meaningless text, e.g., a random combination of several words.
    % \item \textbf{only text} refers to the scenario where the image stored in memory is arbitrarily associated with text that is randomly selected from a different image also stored in the memory.
    (2) \textbf{Q-I-M-IT}
    % "Query-Image-Memory-Image and Text" 
    describes the standard setting for MUIER by pairing the memory image with its corresponding text label. Specifically, when we use an image as a query, we will calculate its similarity between the query image $i_q$ and the memory image $i_m$ as well as its caption $c_m$, which means the final similarity $S=cos(i_q, i_m) + cos(i_q, c_m)$.
    % (3) \textbf{Q-I-M-IT w/ mask} 
    % "Query-Image-Memory-Image and Text w/ mask" 
    % pertains to the scenario in which a subset of words (N=3) within the label text are removed, thereby only a fraction of the words within the label text accessible to the MLLM during the evaluation phase (to determine if a sentence fragment is sufficient for extracting textual features with less memory requirement). 
    % We replace every sequence of words in the logical form
% that appears in the input utterance with the string
% [MASKED] for both the target utterance and in-context examples
\begin{figure}
    \centering
    \includegraphics[width=0.4\textwidth]{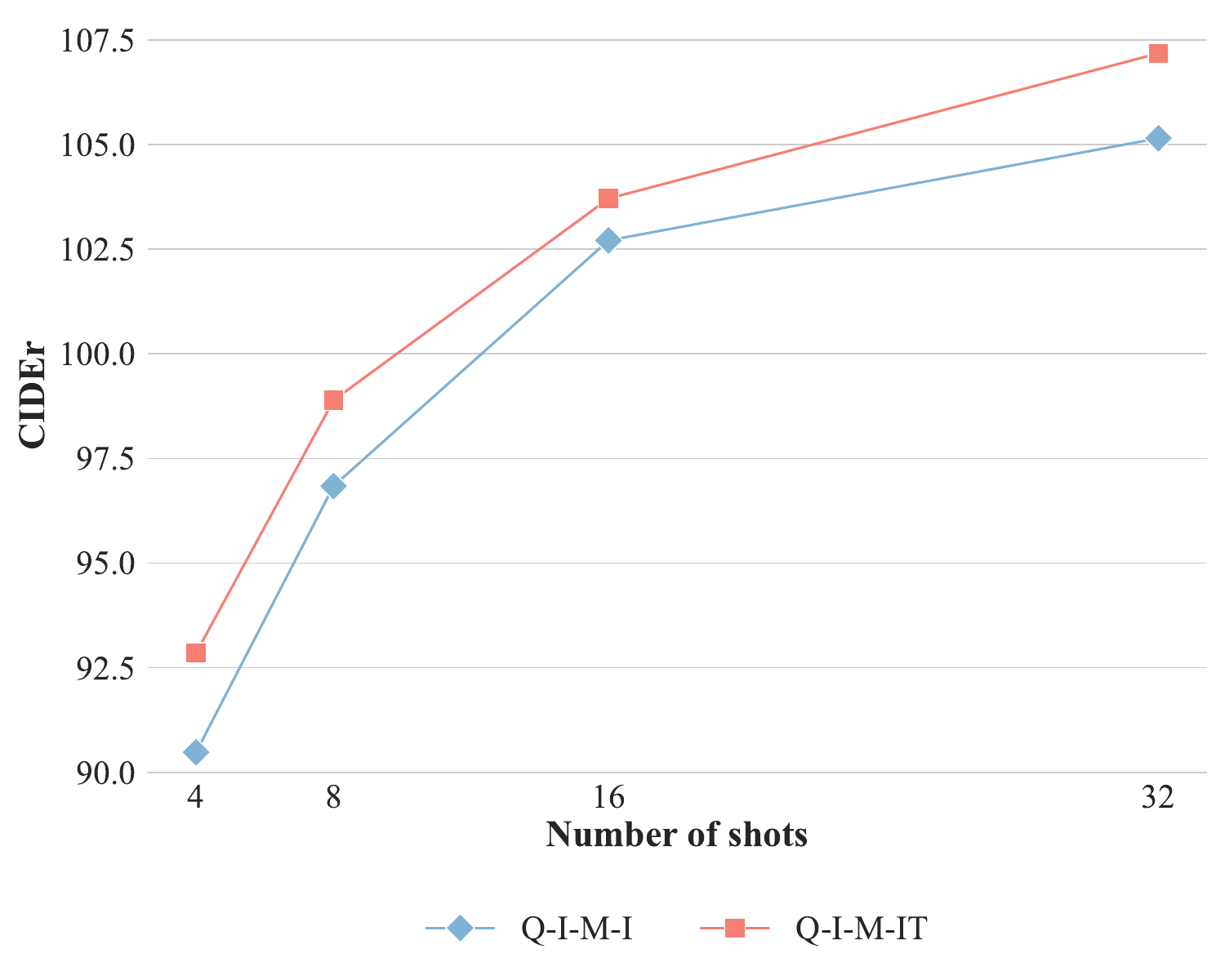}
    \caption{
    % The M-ICL performance is almost the same when removing the visual information in the demonstration. Compared to the standard scenario, exclusion and replacement of images in the demonstration hardly impact the In-Context Learning performance (as shown in the first three bars of each sub-figure). Conversely, the removal of the query image results in substantial performance degradation (as indicated by the last bar in each sub-figure).
    The introduction of textual information in the unsupervised method leads to a higher M-ICL performance across all numbers of shots, demonstrating the importance of text modality. 
    % Compared to the standard scenario (Q-I-M-I), the addition of texts in the retrieval results in substantial performance improvement. 
    % Conversely, the impact of masked texts on the Multimodal In-Context learning performance is minor and decreases as the number of shots is increased.
    }
    \label{fig:coco1}
\end{figure}

Figure \ref{fig:coco1} presents the M-ICL performance in different unsupervised retriever settings, given the same query (Q) and memory (M) configuration. Compared with the standard setting, the Q-I-M-IT setting exerts a positive impact on the M-ICL performance to varying degrees, which verifies the substantial influence of textual content on M-ICL performance.
% and some performances remain relatively unchanged. 
% Specifically, the Q-I-M-IT configuration yields a notable enhancement in Multimodal In-Context Learning (M-ICL) efficacy, manifesting in an increase of nearly 2\% in image captioning performance. Additionally, the performance discrepancy between Q-I-M-IT with mask and Q-I-M-IT persists across all shots but the 16-shot scenario, underscoring the critical importance of comprehensive and precise textual data in the retrieval of in-context examples. In summary, Figure \ref{fig:coco1} verifies the substantial influence of textual content on M-ICL performance.

\subsection{Multimodal Supervised Prompt Retriever}
% Based on the result for \ref{impo}, we utilize both image and textual information for the retrieval of Top-K examples that will be further used for the training of our proposed multimodal prompt retriever. Obviously, it will take a huge cost of time and computing resources if we score each candidate in the memory for MLLM training due to the large size of existing MLLM and the high cost of preprocessing image information. As a result, following {}, we first filter top-k candidates based on the similarity of multimodal information using the CLIP model. Specifically, the cosine similarity was applied here.
The unsupervised methodology depicted earlier is not explicitly designed for multimodal in-context learning applications. Instead, its efficacy relies on the preliminary training phase of the feature extraction mechanism, where the objective function utilized may not align with the demands of multimodal in-context learning scenarios. In contrast, we propose a novel strategy that is based on supervised in-context examples retrieval, assuming the availability of labeled training data as shown in Figure \ref{fig:msier}. The primary aim of this approach is to refine the original retriever to ensure that the chosen in-context example(s) contribute effectively towards enhancing the log-likelihood maximization.

It is important to acknowledge that evaluating each candidate within the dataset for MLLM training entails a significant expenditure of time and computational resources. This is attributed to the extensive size of existing MLLMs and the substantial overhead associated with preprocessing image data. Consequently, in alignment with the methodologies outlined in \cite{rubin2022learning}, we initially narrowed down the Top-$N$ ($N$=50) candidates by assessing the cosine similarity of multimodal information through the application of the unsupervised retriever. 
Drawing upon the findings presented in \ref{impo}, our approach integrates both visual and textual data for the identification of Top-$N$ examples, which are subsequently employed in the training of our proposed multimodal in-context examples retriever. 
% It is important to acknowledge that evaluating each candidate within the dataset for MLLM training entails a significant expenditure of time and computational resources. This is attributed to the extensive size of existing MLLMs and the substantial overhead associated with preprocessing image data. Consequently, in alignment with the methodologies outlined in \cite{rubin2022learning}, we initially narrowed down the Top-$K$ candidates by assessing the cosine similarity of multimodal information through the application of the unsupervised retriever. 

% The unsupervised approach previously described is not specifically tailored for multimodal in-context learning. Rather, it depends on the pre-training of the feature extractor, and the objective function employed during this pre-training might not correspond with the requirements of in-context learning. In contrast, we suggest an alternative method predicated on supervised prompt retrieval, predicated on the presumption that the source data includes labels. The objective of this method is to optimize the score function, denoted as , in a manner that ensures the selected in-context example(s) are conducive to maximizing the log-likelihood.

% \subsection{Training and Evaluation}
\textbf{Scoring} 
% Concerning the scoring of retrieved candidates for each training example by unsupervised retriever, we first construct effective prompts as suggested in , as shown in Table \ref{Tab:prompt}. The scoring MLLM receives a specific prompt as input and makes predictions for the orange part, which is then utilized for the calculation of NLL loss which is considered as the score of this training example. For example, for the image captioning task, we first obtain a query image and its respective context image-text pair. Afterward, the scoring MLLM takes a list of images and the prompt as input and outputs its logits for the answer part. Lastly, generated logits will be used for calculating NLL loss which is the achieved score for this example. As a result, the retrieved candidates will be resorted on the basis of scores for the next formation of positive and negative samples.
In the process of scoring retrieved candidates for each training instance via an unsupervised retriever, we initially devise effective prompts as illustrated in Table \ref{Tab:prompt} following \cite{awadalla2023openflamingo}. The MLLM scorer is furnished with a specific image or image-text prompt as input and proceeds to make predictions for the designated orange segment. These predictions are subsequently leveraged to compute the NLL loss, which serves as the performance metric for the given in-context instance. 
% For example, within the context of image captioning tasks, a query image along with its associated context in the form of an image-text pair is first procured. The MLLM scorer then processes a combination of images and the context prompt, outputting logits for the required response part. These generated logits are utilized to calculate the NLL loss, representing the score for that particular example. 
Consequently, the candidates are reorganized based on their scores to facilitate the subsequent selection of positive and negative samples for contrastive learning.

\begin{table*}
\centering
\begin{tabular}{lll}
\hline
\textbf{Tasks} & \textbf{Prompts}\\
\hline
Image Captioning &  <image> Output: \textcolor{orange}{[caption]}  \\
VQA & <image> Question: [question] Short answer: \textcolor{orange}{[answer]} \\
Rank Classification & <image> is an image with: ‘[text]’ written on it. Is it hateful? Answer: \textcolor{orange}{[answer]} \\
% \citeyearpar{Gusfield:97} & \verb|\citeyearpar| & \verb|\shortcite| \\
\hline
\end{tabular}
\caption{\label{citation-guide}
% Prompts used for the scoring process of MLLM. The score for each candidate in the memory is based on orange parts.
Prompts used for different tasks. Orange parts are used for calculating scores for each candidate in memory.
}
\label{Tab:prompt}
\end{table*}

% apply a pre-trained scoring MLLM

\textbf{Training} 
% Following the contrastive learning paradigm from \cite{rubin-etal-2022-learning}, a query encoder $E_q$ that takes an input in an image-question pair, and a context encoder $E_c$ that receives one candidate prompt in the form of an image-prompt pair are introduced for extracting features that we aim to further optimize. Both encoders are initialized with CLIP V-L-14. At each step, a minibatch $B$ from the training dataset was sampled for the fine-tuning of the CLIP model. Next, for each example in the constructed memory $B$, the corresponding positive example and negative example are sampled from the Top-K positive and negative candidates separately for the goal of learning a similarity metric such that given a test example $x_test$, it will be similar to training examples that lead to decoding of $y_test$. The calculation of contrastive loss is as follows:
Adopting the contrastive learning framework as mentioned by \cite{rubin2022learning}, we introduced a query encoder designed to process inputs comprising image or image-question pairs, and a context encoder tasked with handling candidate prompts represented as image-prompt pairs, for feature extraction aimed at subsequent optimization. Both encoders are initialized with vision encoders and text encoders of CLIP model \cite{clip-radford21a}. During each iteration, a minibatch $B$ is selected from the training dataset to fine-tune the CLIP model. Furthermore, for every instance within the assembled memory $B$ with $N$ image-text pair, both a positive and a negative example are independently sampled from the Top-$K$ positive and negative candidates. This sampling strategy is implemented to develop a similarity metric. Specifically, this metric should ensure that for a given test example $x_q$, it exhibits similarity to training instances that facilitate the decoding of $y_q$. The formulation of the contrastive loss is executed as follows:

\begin{equation}
    \mathcal{L} = -\log{\frac{e^{cos(x_q, e^+)}}
    {e^{cos(x_q, e^+)}+ 
    \sum \limits_{i=1}^{2N-1} e^{cos(x_q, e^-_i)}}}
\end{equation}

where cos(·, ·) measures the consine similarity. $e^+$ denotes the feature representation of a positive example, and $e^-$ denotes the feature representation of a negative example.
% It is worth noting that for mini-batch training, the negative set N contains a negative example of $e_n$ sampled from the Top-$K$ negative set and other examples within the same mini-batch.

% \textbf{Evaluation} 

\subsection{Importance of Textual Information in Supervised Retriever}
\label{investigate}
In Section \ref{impo}, the significance of various modalities is examined within the context of unsupervised retrieval, yet the impact of textual data on MSIER has not been investigated. To thoroughly assess the contribution of text in the training of supervised retrievers, we conducted a detailed comparison of two scenarios involving fine-tuned CLIP models, distinguishing them based on their approach to incorporating textual information within the image captioning task.

\begin{figure}
    \centering
    \includegraphics[width=0.45\textwidth]{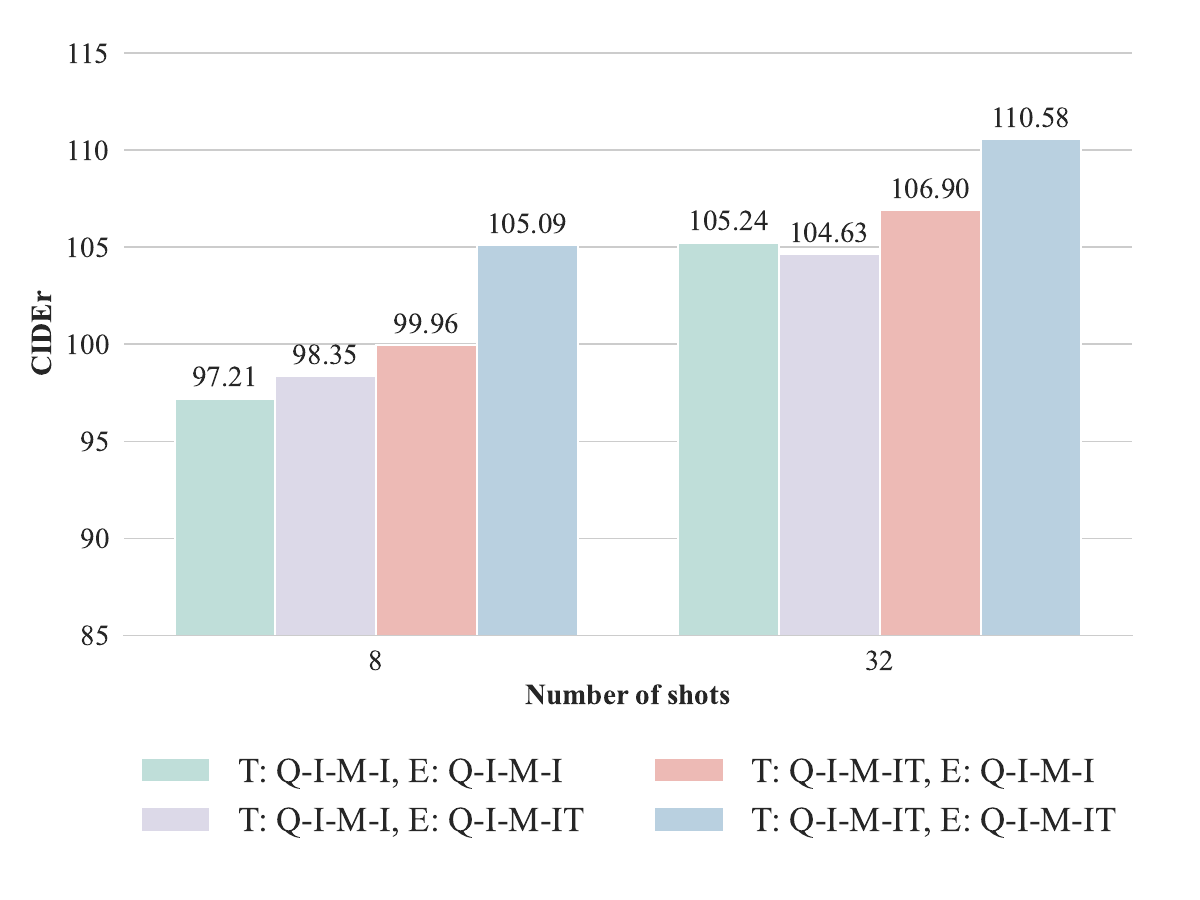}
    \caption{Impact of texts on proposed MSIER method. `\textbf{T}' denotes the \textbf{T}raining setting and `\textbf{E}' denotes the \textbf{E}valuation setting.}
    \label{fig:T-text}
\end{figure}

We decoupled the training and evaluation settings to investigate the significance of textual data in our proposed supervised retrieval framework. Specifically, we employed distinct query-memory configurations during the retriever's training phase and evaluation phase. As depicted in Figure \ref{fig:T-text}, two training configurations -- \textbf{Q-I-M-I} and \textbf{Q-I-M-IT} -- were implemented, and their performance was assessed under two corresponding settings to ensure an equitable comparison. Within the context of the Q-I-M-I setting, textual data during the training phase has a negligible influence on the retriever's performance. In contrast, incorporating textual information during the training of the supervised multimodal retriever markedly enhances its effectiveness, resulting in a significant performance improvement.

\section{Experiment}
\label{experiment}

% In this section, a comprehensive evaluation is conducted employing various prompt selection methodologies as delineated in Section 3.1, in addition to comparing their resilience against distribution shifts as outlined in Section 3.2. Furthermore, extensive quantitative and qualitative analyses are presented in Section 3.3 to elucidate the mechanisms underpinning the efficacy of our methods and to facilitate their more effective implementation in practical scenarios. The source code will be made available to the community to enable the replication of the complete experiments.

\subsection{Datasets}
Following \cite{alayrac2022flamingo}, we focus on three representative multimodal tasks and the details about the datasets used for these tasks
as follows. MS COCO \cite{lin2015microsoft} for image captioning, OK-VQA \cite{marino2019okvqa} for Visual Question Answering, and HatefulMemes \cite{kiela2021hateful} for rank classification.
Accuracy on the test split is measured for OK-VQA. In MSCOCO, evaluation utilizes CIDEr scores \cite{vedantam2015cider} on the Karpathy-test split. For HatefulMemes, the AUC ROC is calculated. All results are averaged with three runs. For further details of these downstream tasks and corresponding datasets, please refer to Appendix~\ref{sec:appendix::downstream} and \ref{sec:appendix::datasets}. The experiments conducted herein utilize the OpenFlamingo-3B framework \cite{awadalla2023openflamingo}. Comprehensive details regarding OpenFlamingo can be found in Appendix~\ref{sec:appendix::backbone}.

\subsection{Compared Methods} 
% The comparative analysis primarily focuses on the efficacy of several methodologies: (1) The baseline approach, denoted as 'Random,' which involves the indiscriminate selection of in-context examples from the originating training dataset; (2) The RICES (Retrieval-based In-Context Example
% Selection) method that finishes unsupervised retrieval based on image similarity between the query and in-context examples from memory, termed as 'Q-I-M-I'; (3) Q-T-M-T, a variant of 'Q-I-M-I', selected demonstrations based on text similarity; (4)
% The initial method proposed by our study, Multimodal Unsupervised Prompt Retrieval (MUPR), leverages readily available features for the purpose of identifying the nearest examples through search on the basis of different mulmodality. This method's foundation is the CLIP vision and text encoder, as developed by Radford et al. (2021), which underwent pre-training via multimodal contrastive learning techniques; (5) Multimodal Supervised Prompt Retrieval (MSPR), delineates our secondary proposition, which entails the refinement of CLIP's two encoders. This refinement process is achieved through direct optimization aimed at enhancing in-context learning outcomes within subsequent datasets. The supervised model is trained for 30 epochs using AdamW. The initial
% learning rate is set to 0.00001, decayed by the cosine annealing rule.

The analysis evaluates various methodologies, including the Random baseline, RICES (Q-I-M-I), Q-T-M-T textual variant, Multimodal Unsupervised In-context Examples Retrieval (MUIER) using all multimodality aspects, and the proposed approach:  Multimodal Supervised In-context Examples Retrieval (\textbf{MSIER}) enhancing CLIP's dual encoders. MSIER undergoes 30 epochs of training with the AdamW optimizer \cite{kingma2017adam}, warming up to a peak learning rate 1e-5, subject to reduction using the cosine annealing rule. For further details of these methods, please refer to Appendix~\ref{sec:appendix::baselines}.

\begin{table*}[ht]
\centering
\resizebox{\textwidth}{!}{%
    \begin{tabular}{ c c c c c c c c c c c c c c c }
      \toprule
      & \multicolumn{4}{c}{\textbf{MS COCO}} & \multicolumn{5}{c}{\textbf{OK-VQA}} & \multicolumn{5}{c}{\textbf{HatefulMemes}} \\
      \cmidrule(l){2-5} 
      \cmidrule(l){6-10} 
      \cmidrule(l){11-15} 
      \textbf{} & \textbf{Random} & \textbf{RICES} & \textbf{MUIER} & \textbf{MSIER} & \textbf{Random} & \textbf{RICES} & \textbf{Q-T-M-T} & \textbf{MUIER} & \textbf{MSIER} &\textbf{Random} & \textbf{RICES} & \textbf{Q-T-M-T} & \textbf{MUIER} & \textbf{MSIER}\\
      \midrule
      4 shots  & 77.19 & 91.43 & 92.78 & \textbf{100.58} & 30.32 & 31.09 & 30.66 & 32.46 & \textbf{33.64} & 50.56 & 65.74  &  63.05 & 68.78 & \textbf{70.77} \\
      8 shots & 85.97  & 96.91 & 98.93 & \textbf{105.09} & 31.08 & 32.36 & 31.90 & 34.82 & \textbf{36.16} & 52.67 & 67.28  &  63.64 & 70.13 & \textbf{72.53} \\
      16 shots & 90.51	& 102.22 &  103.65 & \textbf{108.41}  & 30.93 & 33.58 & 32.57 & 36.04 & \textbf{37.12} & 48.45 & 68.41  &  62.55 & 70.34 & \textbf{72.59} \\
      32 shots  & 92.47 & 106.30 & 107.24 & \textbf{110.58}  & 31.02 & 35.41 & 33.32 & 36.79 & \textbf{38.35} & 50.91 &  68.36  &  63.27 & 71.50 & \textbf{73.80} \\
      Avg  & 86.54  &  99.22 & 100.65 & \textbf{106.17}  & 30.84  & 33.11 & 32.12 & 35.03 & \textbf{36.32} & 50.65 & 67.45 &  63.13 & 70.19 & \textbf{72.42}\\
      \bottomrule
    \end{tabular}
}
\caption{Results of M-ICL performance of random selection, RICES, MUIER, and MSIER method on MS COCO, OK-VQA, and HatefulMemes dataset.}
\label{tab:all}
\end{table*}

\subsection{Main Results}
As shown in Table \ref{tab:all}, compared with Random and RICES, MUIER method shows further enhancements, underscoring the benefits of integrating images and text in a retrieval strategy without explicit supervision. This approach enables the model to better understand in-context examples, with the textual content offering additional direction, thus improving the retrieval of more relevant examples. MSIER method outperforms all with a significant rise in performance, evidenced by a $5.52$ increase in the CIDEr score of MS COCO dataset. This suggests that a refined retrieval mechanism, informed by the foundational knowledge of multimodal large language models (MLLMs) and tailored to select correct image-text pairings, yields superior learning outcomes. Furthermore, MSIER with only 4 shots outperforms random selection with 32 shots. Thus, employing our method can enable MLLMs to achieve comparable M-ICL performance with limited memory usage.
\begin{figure}
    \centering
    \includegraphics[width=0.5\textwidth]{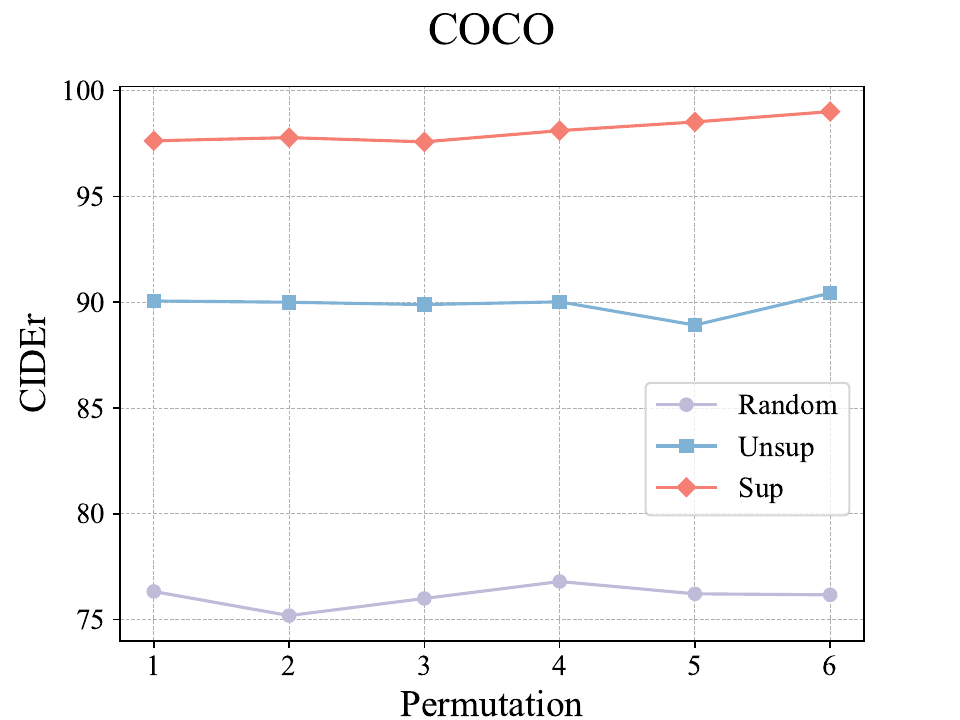}
    \caption{ Impact of the order of retrieved multimodal in-context examples.}
    \label{fig:order}
\end{figure}
\subsection{Further Analysis}
\begin{figure*}
    \centering
    \includegraphics[width=0.7\textwidth]{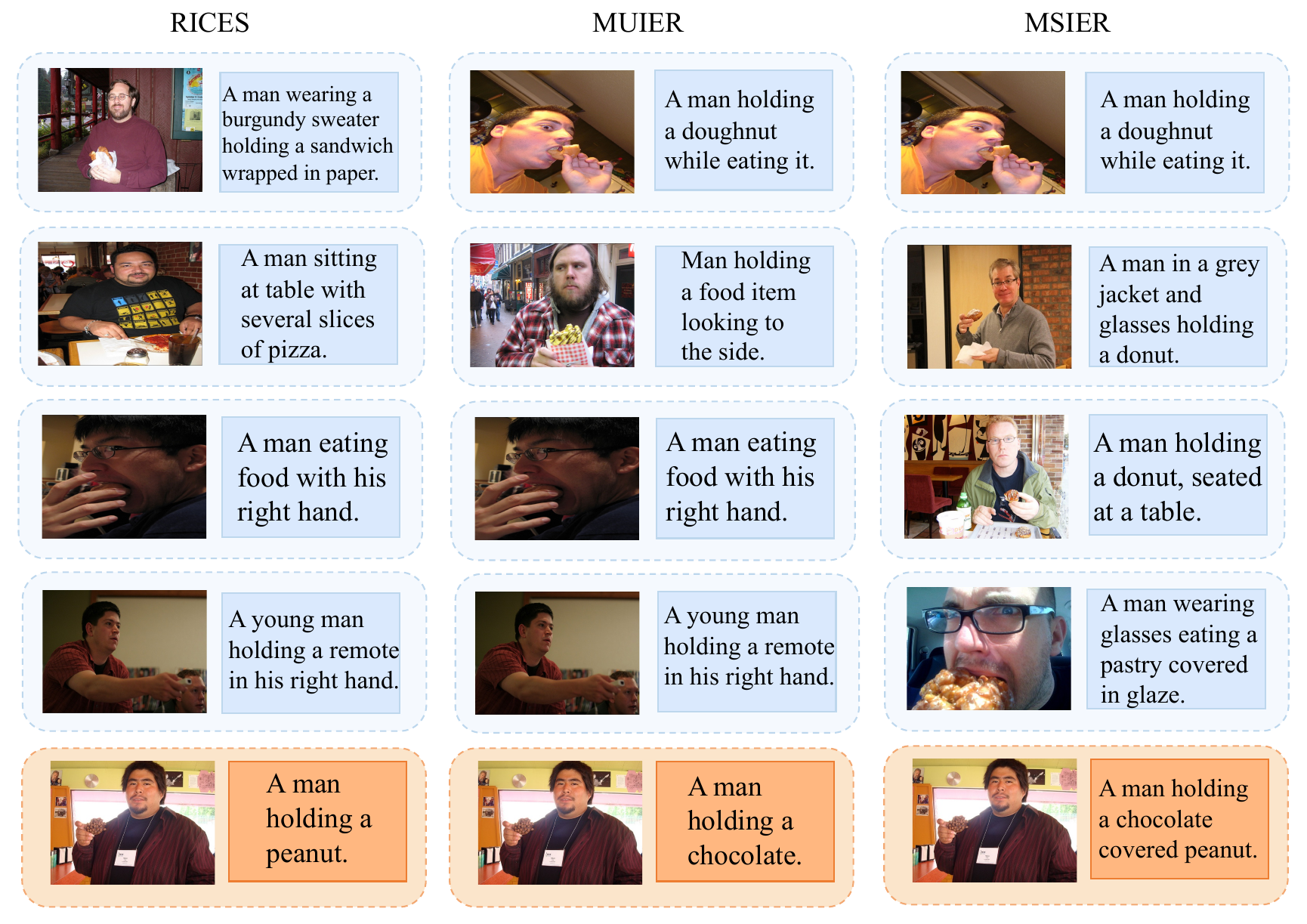}
    \caption{Multimodal in-context examples retrieved by RICES, MUIER and MSIER. In each grid, the first four blue rows contain the prompt while the orange row contains the query image and prediction. Further outcomes are detailed in the Appendix~\ref{sec:appendix::examples}.}
    \label{fig:check}
\end{figure*}
\textbf{How does the MSIER improve M-ICL specifically?}
To address this query, we employed a multimodal analysis of the in-context examples identified by MUIER and MSIER as depicted in Figure \ref{fig:check}. Our examination concentrates on image captioning by selecting one example from the MS COCO dataset where the number of shots is 4. Concretely, the columns from left to right delineate the retrieved in-context example with RICES, MUIER, and MSIER separately. The blue rows are the retrieved multimodal in-context examples,  that is, a pair comprising input and output, whereas the subsequent orange rows present the query image alongside the model's prediction. 
% Through a comparative analysis of the in-context examples selected by RICES, MUIER, and MSIER, we elucidate the superior performance of MSIER over MUIER; the examples identified by MSIER exhibit a closer semantic similarity to the queries. 
In the given figure, examples identified by MSIER exhibit a closer semantic similarity to the queries than RICES and MUIER. Consequently, the MLLM made a more accurate prediction that includes the caption "a chocolate covered peanut", which is not captured by previous predictions using RICES and MUIER methods for retrieving in-context examples. 

% This pattern of similarity is also observed across other categories and tasks, for which readers are directed to consult the \textcolor{red}{appendix}.

\textbf{How does the order of in-context examples affect M-ICL?} To understand if changing
the order of multimodal in-context examples makes a difference, we
fixed the number of in-context examples to 3, and evaluate all
possible permutations. As shown in Figure \ref{fig:order}, the standard deviation is
generally small, so the order is not a concern as long as
high-quality examples are chosen.

% \begin{table}[ht]
% \caption{Impact of the order of supervised retrieved demonstrations.}
% \label{tab:order}
% \resizebox{0.5\textwidth}{!}{%
% \centering
% \begin{tabular}{ c c c c c c }
%   \toprule
%   \textbf{Methods} & \textbf{4 shots} & \textbf{8 shots} & \textbf{16 shots} & \textbf{32 shots} & Avg\\
%   \midrule
%   Random  & 50.6 & 52.0 & 48.5 & 50.2 & \\
%   Unsup (Q-I-M-IT) & 68.78 & 70.13 &  70.34 & 71.5  &  \\
%   Sup (Q-I-M-IT) & 70.77 & 72.53 & 72.59 &73.8  & \\
%   \bottomrule
% \end{tabular}
% }
% \end{table}

\textbf{How is the transferability of proposed MSIER?}
The compositional characteristics of natural language and images are general, meaning the retriever may exploit similar knowledge in different tasks or scoring MLLMs. This motivates
us to explore whether the proposed MSIER trained on one dataset or MLLM scorer can be directly transferred to others
without further tuning. This is a practical research question
as training a retriever for each dataset or MLLM scorer can
be costly in real applications.

\begin{itemize}
    \item \textbf{Datasets} To measure the transferability of supervised retrievers between datasets, we evaluated the performance of MSIER trained on OK-VQA and HatefulMemes datasets on MS COCO dataset with different numbers of shots of retrieved in-context examples. The result in Table \ref{tab:dstaset} manifests some transferability of our proposed MSIER between multiple multimodal datasets. Enhanced performance of MSIER with training on the OK-VQA dataset suggests that data form and volume significantly influence MSIER's effectiveness.
    \item \textbf{MLLM Scorer} Regarding the impact of MLLM scorer on the transferability of MSIER, our methodology involves deploying a retriever trained utilizing the OpenFlamingo-3B model as the scoring model, subsequently applied to the inference processes of the OpenFlamingo-9B model. The outcomes in Table \ref{tab:scorer} demonstrate that the MSIER approach, when utilizing the 3B model as a scorer, manifests superior transferability and outperforms the MUIER-9B method. This implies that supervised retrieval methods, initially trained on smaller-scale models, retain effectiveness upon application to larger models, obviating the necessity for separate training for larger models and thus enhancing the cost-efficiency of inference processes in larger-scale models.
\end{itemize}

\begin{table}[ht]
\centering
\resizebox{0.48\textwidth}{!}{%
    \begin{tabular}{ c c c c c c }
      \toprule
      \textbf{Method} & \textbf{4 shots} & \textbf{8 shots} & \textbf{16 shots} & \textbf{32 shots} & \textbf{Avg}\\
      \midrule
      MUIER (Q-I-M-IT) & 92.78	& 98.93 &  103.65 & 107.24  & 100.65 \\
      \midrule
      HMM (Q-I-M-IT) & 91.92	& 97.87	& 103.82 &  107.30 & 100.23\\
      OKVQA (Q-I-M-IT) & 93.28 &	100.72	& 105.95 & 109.02  & 102.24 \\
      \bottomrule
    \end{tabular}
}
\caption{M-ICL performance of MSIER trained using HMM and OK-VQA datasets on MS COCO dataset.}
\label{tab:dstaset}
\end{table}

\begin{table}[ht]
\centering
\resizebox{0.48\textwidth}{!}{%
    \begin{tabular}{ c c c c c c }
      \toprule
      \textbf{Method} & \textbf{4 shots} & \textbf{8 shots} & \textbf{16 shots} & \textbf{32 shots} & \textbf{Avg}\\
      \midrule
      Random  & 89.03 & 96.30 & 98.78 & 99.91 &  96.01\\
      RICES (Q-I-M-I) & 93.28 & 99.89 & 105.71 & 108.79 & 101.92 \\
      \midrule
      MUIER-9B (Q-I-M-IT) & 94.53 & 100.48 & 106.11 & 109.95  & 102.77 \\
      MSIER-3B (Q-I-M-IT) & 100.73 & 103.93 & 107.50 & 110.60 & 105.69 \\
      \bottomrule
    \end{tabular}
}
\caption{Results of M-ICL performance of random selection, RICES, and MUIER-9B and MSIER-3B method on MS COCO dataset using OpenFlamingo-9B for inference.}
\label{tab:scorer}
\end{table}

\subsection{Ablation Study}
This section presents an extensive ablation study based on the OKVQA and MS COCO datasets to verify the best setting for our proposed model’s components.
% following existing works [2], [3], [13], [64] in the early activity prediction community.

\textbf{Impact of Different Modality Encoders in CLIP}
Unlike ICL retrieval in NLP and CV, M-ICL retrieval employs multiple encoders in its backbone retriever to extract features from different modalities. To investigate the roles of the text and image encoders within the base CLIP model in our proposed MSIER framework, we selectively freeze one encoder during MSIER training. Table \ref{tab:EF} demonstrates that the update on image encoder has a more significant effect on M-ICL retrieval performance improvement because freezing the image encoder results in a notable performance drop compared to the original MSIER (1.39 $\downarrow$). In contrast, freezing only the text encoder yields similar or even improved retrieval performance, suggesting that the original trained MSIER may have overfitted to some degree on the text encoder.
\begin{table}[ht]
\centering
\resizebox{0.48\textwidth}{!}{%
    \begin{tabular}{ c c c c c c }
      \toprule
      \textbf{Method} & \textbf{4 shots} & \textbf{8 shots} & \textbf{16 shots} & \textbf{32 shots} & \textbf{Avg}\\
      \midrule
      MUIER (CLIP)  & 92.78 &	98.93	& 103.65 &	107.24 & 100.65 \\
      MSIER (CLIP) & 100.58 &	105.09	& 108.41 &	110.58 & 106.17 \\
      \midrule
      MSIER (CLIP-Freeze T) & 100.82 & 105.90	& 109.17 &	110.24 & 106.53 \\
      MSIER (CLIP-Freeze I) & 98.13 &	103.37	& 107.59 &	110.01 & 104.78 \\
      \bottomrule
    \end{tabular}
}
\caption{Comparison of M-ICL performance of  MSIER method on MS COCO dataset with freezing different modality encoders of the backbone retriever.}
\label{tab:EF}
\end{table}

\textbf{Impact of Number of Candidates}
We proceed to assess the performance of MSIER across a range of candidate numbers. In MSIER, we passed the candidates to a scoring LM and label the top-$K$ and the bottom-$K$ as positive and negative examples respectively. If the number of candidates is not large enough, it will not hold sufficient information for contrastive learning; conversely, if we set the bank with a large size, it will contain a quantity of irrelevant items. 
The results in Table \ref{tab:INC} present $K$=5 as the best choice on OK-VQA dataset.
% A representation of the results can be found in Fig. 8. The AUC, a metric quantifying average precision across all observation ratios, exhibits a rapid increase from a small to a large bank size, before stabilizing at a sufficiently large size (e.g., size 5000). In this case, the additional performance gain is restricted when the inherent threshold is reached, due to the total number of hard pairs in the dataset.

\begin{table}[ht]
\centering
\resizebox{0.47\textwidth}{!}{%
    \begin{tabular}{ c c c c c c }
      \toprule
      \textbf{Method} & \textbf{4 shots} & \textbf{8 shots} & \textbf{16 shots} & \textbf{32 shots} & \textbf{Avg}\\
      \midrule
      K = 1  & 33.57 & 35.93 & 36.92 & 38.07 & 36.12\\
      K = 5  & 33.64 & 36.16 & 37.12 & 38.35 & 36.32 \\
      K = 10 & 33.74 & 35.58 & 36.85 & 37.94 & 36.03 \\
      \bottomrule
    \end{tabular}
}
\caption{Comparison of M-ICL performance of different settings of $K$ for MSIER on OK-VQA dataset.}
\label{tab:INC}
\end{table}

\textbf{Impact of Textual Information in Retrieved In-context Examples}
The role of textual information during the evaluation of M-ICL remains unexplored for our proposed MSIER. In Table \ref{tab:IT}, we find that compared to the standard scenario, the replacement of texts in the in-context examples hugely impacts the M-ICL performance. To be specific, employing a mask on half of the captions through the random selection method results in a significant decrease in the CIDEr score. Correspondingly, the side effect of the removal of texts decreases from MUIER method to MSIER method compared with random retrieval.
% Further, the removal of the query image results in substantial performance degradation.

\begin{table}[ht]
\centering
\resizebox{0.48\textwidth}{!}{%
    \begin{tabular}{ c c c c c c }
      \toprule
      \textbf{Method} & \textbf{4 shots} & \textbf{8 shots} & \textbf{16 shots} & \textbf{32 shots} & \textbf{Avg}\\
      \midrule
      Random  & 77.19 & 85.97 & 90.51 & 92.47 & 86.54 \\
      Random w/ mask  & 2.27 & 2.28 & 2.40 & 2.79 & 2.44 \\
      \midrule
      MUIER (Q-I-M-IT) & 92.78 & 98.93 & 103.65 & 107.24 & 100.65 \\
      MUIER (Q-I-M-IT) w/ mask & 70.32 & 74.34 & 77.68 & 78.62 & 75.24 \\
      \midrule
      MSIER (Q-I-M-IT) & 100.58 & 105.09 & 108.41 & 110.58  & 106.17 \\
      MSIER (Q-I-M-IT) w/mask & 77.62 & 82.00 & 85.83 & 86.77  & 83.06 \\
      \bottomrule
    \end{tabular}
}
\caption{Comparison of M-ICL performance of random selection, MUIER, and MSIER method with masked text in-context examples on MS COCO dataset.}
\label{tab:IT}
\end{table}

\textbf{Impact of Backbone Retriever}
To investigate whether using a different backbone model could improve performance, we further evaluated retrieval methods, MUIER and MSIER, on the image captioning benchmark by utilizing an alternative backbone: EVA-02-CLIP \cite{sun2023evaclip}, a model enhances the original CLIP model with advanced pre-training techniques. The results, reported in Table \ref{tab:evaclip}, show that MUIER and MSIER methods employing EVA-02-CLIP(-L/14) as the backbone retriever demonstrated better performance. This suggests that the EVA-02-CLIP model offers retrieval benefits for the M-ICL of the OpenFlamingo model, with improved feature extraction for both textual and visual information. Furthermore, the proposed MSIER method exhibited a corresponding performance improvement when using the enhanced retriever, explicitly validating MSIER's excellent transferability and ability to achieve better results as multimodal encoders advance. Additional experiments can be found in Appendix \ref{sec:appendix::ibr}.
% This is due to OpenFlamingo's use of the CLIP model as its vision encoder, resulting in poor contextual information from examples retrieved primarily by the ALIGN model, leading to little enhancement in M-ICL performance.

% \begin{table}[ht]
% \resizebox{0.45\textwidth}{!}{%
% \centering
%     \begin{tabular}{ c c c c c c }
%       \toprule
%       \textbf{Method} & \textbf{4 shots} & \textbf{8 shots} & \textbf{16 shots} & \textbf{32 shots} & \textbf{Avg}\\
%       \midrule
%       Random & 77.02 & 84.13 & 88.96 & 92.32 & 85.61\\
%       MUIER (Q-I-M-IT)  & 78.78&	84.1	& 89.22 &	92.07 & 86.04 \\
%       MSIER (Q-I-M-IT) & 78.17 &	83.79	& 89.38 &	94.02 & 86.34 \\
%       \bottomrule
%     \end{tabular}
% }
% \caption{Comparison of M-ICL performance of random selection, MUIER,  and MSIER method on MS COCO dataset with ALIGN model as the backbone retriever.}
% \label{tab:align}
% \end{table}

\begin{table}[ht]
\centering
\resizebox{0.48\textwidth}{!}{%
    \begin{tabular}{ c c c c c c }
      \toprule
      \textbf{Method} & \textbf{4 shots} & \textbf{8 shots} & \textbf{16 shots} & \textbf{32 shots} & \textbf{Avg}\\
      \midrule
      MUIER (CLIP)  & 92.78 &	98.93	& 103.65 &	107.24 & 100.65 \\
      MSIER (CLIP) & 100.58 &	105.09	& 108.41 &	110.58 & 106.17 \\
      \midrule
      MUIER (EVA-02-CLIP)  & 94.02 &	101.70	& 107.06 &	109.71 & 103.12 \\
      MSIER (EVA-02-CLIP) & 104.20 &	107.44	& 111.61 &	113.07 & 109.08 \\
      \bottomrule
    \end{tabular}
}
\caption{Comparison of M-ICL performance of  MUIER and MSIER method on MS COCO dataset with EVA-CLIP2 model as the backbone retriever.}
\label{tab:evaclip}
\end{table}
% Although these three backbones perform differently on image recognition under the
% fine-tuning setting—EVA performed the best—the gap between them for both UnsupPR and SupPR is less than 1%.
% Therefore, we can conclude that the backbone for visual

\section{Conclusion}
\label{conclusion}

In this study, we conducted an extensive evaluation of textual information's role in unsupervised and supervised in-context example retrieval for multimodal in-context learning. We introduce a novel MSIER methodology that incorporates MLLMs' self-contained information to train a supervised prompt retriever. Our experiments across three multimodal tasks demonstrate that integrating text modality and the foundational knowledge of MLLMs significantly enhances the efficiency of example selection, yielding substantial improvements over existing approaches. Moreover, MSIER demonstrates high efficiency by outperforming benchmarks with far fewer examples, while also exhibiting strong transferability by remaining effective for larger models after training on smaller ones, thereby improving cost-efficiency for large-scale inference. Future research could delve into optimizing inter-modal interactions, leveraging our findings for more effective multimodal in-context learning retrieval strategies.

\section{Limitations}
Our research focuses exclusively on the integration of visual and textual data for multimodal in-context learning. Given the expanding use of additional modalities (such as video and audio) in this domain, the development of a comprehensive framework capable of consolidating these varied modalities into a unified representation is increasingly imperative. 

% This work mainly discusses the role of textual information in the retrieval of M-ICL and the efficiency of the proposed supervised prompt retriever, instead of achieving SOTA in MLLM downstream tasks. because in-context learning is a widespread paradigm to stimulate MLLM capabilities for downstream tasks but the final performance (e.g., CIDEr score) is still limited by the foundation MLLM (OpenFlamingo 3B in our experiments).

We observed that enhancements in multimodal in-context learning (M-ICL) performance on the MS COCO dataset surpass those on other datasets. This discrepancy may stem from the necessity for more meticulously constructed prompts in visual question answering (VQA) and rank classification tasks, which demand more than mere questions or captions for effective in-context example retrieval.

Moreover, both diversity and relevance play critical roles in the selection of in-context examples for M-ICL. Future investigations might, therefore, benefit from examining how the relationships between retrieved examples can inform the design of an improved supervised retriever for M-ICL.

\section*{Acknowledgements}
Yang You's research group is being sponsored by NUS startup grant (Presidential Young Professorship), Singapore MOE Tier-1 grant, ByteDance grant, ARCTIC grant, SMI grant and Alibaba grant.

\bibliography{anthology,custom}
\bibliographystyle{acl_natbib}

\newpage
\appendix

\section{More Experimental Details}
\label{sec:appendix}

\subsection{Compared Methods}
\label{sec:appendix::baselines}
The comparative analysis predominantly concentrates on the effectiveness of various methodologies as follows:
\begin{itemize}
    \item The baseline methodology, referred to as \textbf{Random}, encompasses the arbitrary selection of in-context examples from the originating training dataset.
    \item The \textbf{RICES} (Retrieval-based In-Context Example Selection) strategy completes unsupervised retrieval predicated on the image similarity between the query and in-context examples from memory, denominated as Q-I-M-I.
    \item The \textbf{Q-T-M-T} variant, an adaptation of Q-I-M-I, elects in-context examples predicated on only textual similarity.
    \item The standard unsupervised approach used in this research, termed Multimodal Unsupervised In-context Examples Retrieval (\textbf{MUIER}), employs readily accessible features to identify the most similar examples via search, based on all multimodality aspects. 
    \item Multimodal Supervised In-context Examples Retrieval (\textbf{MSIER}) represents our secondary proposal, which involves the enhancement of CLIP's dual encoders. CLIP ViT-L-14 is applied for all experiments. This enhancement process is facilitated through direct optimization to improve M-ICL performance. The supervised model undergoes training for 30 epochs utilizing the AdamW optimizer. The initial learning rate is established at 0.00001, subject to reduction according to the cosine annealing rule.
\end{itemize}

\subsection{Multimodal Large Language Model}
\label{sec:appendix::backbone}
The architecture of OpenFlamingo encompasses a fixed large language model featuring a decoder-only configuration (for instance, MPT [31]), accompanied by a static visual encoder (such as CLIP-ViT \cite{clip-radford21a}), which is succeeded by a trainable perceiver resampler. To facilitate the integration of visual and linguistic data, trainable cross-attention layers are strategically interspersed among the pre-trained language model layers. OpenFlamingo underwent pre-training on the LAION-2B \cite{schuhmann2022laion5b} and Multimodal C4 \cite{zhu2023multimodal} datasets and its performance is on par with that of Flamingo \cite{alayrac2022flamingo}, demonstrating its competitive edge. 

Some well-known MLLMs such as LLaVA and Minigpt-4 do not discuss their in-context learning ability or present any experimental results on in-context learning in their papers. In contrast, the Flamingo model is one of the first and most renowned MLLMs to include experiments on in-context learning in its paper. Since the original Flamingo model is not open-sourced, we opt for the open-source version, OpenFlamingo, as the backbone. We use its in-context learning performance on three representative downstream tasks in its paper for a fair comparison.

% \subsection{Illustration of retrieved in-context examples on VQA task}
% In the supplementary material, we illustrate more multimodal in-context
% learning results of VQA in Figure .

\subsection{Downstream Tasks}
\label{sec:appendix::downstream}
\begin{itemize}
\item \textbf{Image Captioning} is a task where a model generates a textual description of an image. It combines elements of computer vision and natural language processing to interpret the contents of an image and articulate them in human language. The challenge is recognizing the objects within the image and describing their attributes and the relations between them in a coherent sentence structure.
\item \textbf{Visual Question Answering (VQA)} In this task, a model is given an image along with a natural language question about the image, and it must provide an appropriate answer. To provide an appropriate answer, the model must understand both the visual content of the image and the semantics of the question text. It often involves aspects of object detection, attribute recognition, and language understanding to correctly answer questions such as "What color is the car?" or "How many people are in the room?"
\item \textbf{Rank Classification} is a task that involves ordering or prioritizing a set of items or entities based on specified criteria, which may be derived from multiple input modalities like text and images. For example, in a multimodal setting, this could involve analyzing text and images together to rank products in an e-commerce setting based on relevance to a search query or user preference. More broadly, rank classification could involve any task where items need to be ordered or prioritized, and it can be extended to scenarios like sentiment analysis, where responses are classified into ranked categories of sentiment intensity.
\end{itemize}

\subsection{Datasets}
\label{sec:appendix::datasets}
A detailed explanation of used datasets corresponding to evaluated downstream tasks is provided:
\begin{itemize}
\item \textbf{MSCOCO} is a large-scale dataset for object detection, segmentation, and captioning. It provides diverse images with complex scenes and multiple objects in context. Annotations include object segmentation, recognition, and image captions. The dataset contains approximately 330K images, with 1.5 million labeled instances across 80 object categories. We utilize Karparthy's split: training (83K images), validation (5K images), and test (5K images) sets.

\item \textbf{OK-VQA} is a dataset designed for open-ended Visual Question Answering that requires external knowledge beyond image content. It features questions demanding multi-modal knowledge, combining visual cues with general world knowledge. The dataset comprises over 14,000 images sourced from the MSCOCO dataset, with around 14,000 questions spanning 10 categories. We leverage its structured format, which includes a balanced mix of 9,009 training and 5,046 testing questions, to assess the capability of models to integrate visual understanding with external knowledge sources.

\item \textbf{HatefulMemes} A dataset constructed for the detection and classification of hate speech in multimodal content. It uniquely combines textual and visual elements to challenge models in understanding complex, nuanced expressions of hate speech. The dataset consists of 10,000+ meme images, annotated for hate speech detection, with a distribution of 8,500 for training and 3,000 for testing. The Hateful Memes dataset provides a significant challenge in discerning subtle contextual cues and cultural references, requiring advanced multimodal analysis capabilities.
\end{itemize}

\subsection{Retrieval Methods}
In our main experiments, MUIER and MSIER utilized all modalities to calculate similarity scores and directly selected the most similar items as in-context examples. Recently, some newly proposed retrieval methods have emerged, such as Mixed Modality In-Context Example Selection \cite{chen2023understanding}, which uses visual features to retrieve a set of top-N candidate examples and then employs textual information to retrieve the final in-context examples (with varying numbers of examples, such as 4, 8, 16, or 32). We also validated the efficiency of MSIER in this scenario, and Table \ref{tab:mmices} shows that, although we did not train a new MSIER specifically designed for the new retrieval method, it still demonstrated better performance on the VQA task. This further highlights the generality of our proposed MSIER framework.

\begin{table}[ht]
\centering
\resizebox{0.48\textwidth}{!}{%
    \begin{tabular}{ c c c c c c }
      \toprule
      \textbf{Method} & \textbf{4 shots} & \textbf{8 shots} & \textbf{16 shots} & \textbf{32 shots} & \textbf{Avg}\\
      \midrule
      MMICES-CLIP & 32.96 & 33.60 & 35.89 & 36.13 & 34.65 \\
      MMICES-MSIER &33.08 & 34.36 & 35.92 & 36.84 & 35.05  \\
      \bottomrule
    \end{tabular}
}
\caption{Comparison of M-ICL performance of MMICES with CLIP as retriever, and MMICES with MSIER as retriever on MS COCO dataset. We chose the top 50 candidates for further selection of in-context examples in this experiment.}
\label{tab:mmices}
\end{table}

\textbf{Impact of Backbone Retriever}
\label{sec:appendix::ibr}
We further evaluated the retrieval methods MUIER and MSIER using a different backbone: ALIGN \cite{jia2021scaling}, a model pre-trained on noisy image-text pairs using contrastive learning. The results, reported in Table \ref{tab:align}, show that the MUIER and MSIER methods did not demonstrate explicit advantages over random selection. This suggests that the ALIGN model does not offer retrieval benefits for the M-ICL of the OpenFlamingo model. This can be attributed to the fact that OpenFlamingo utilizes the CLIP model as its vision encoder, resulting in poor contextual information from examples primarily retrieved by the ALIGN model, leading to little enhancement in M-ICL performance.

\begin{table}[ht]
\resizebox{0.48\textwidth}{!}{%
\centering
    \begin{tabular}{ c c c c c c }
      \toprule
      \textbf{Method} & \textbf{4 shots} & \textbf{8 shots} & \textbf{16 shots} & \textbf{32 shots} & \textbf{Avg}\\
      \midrule
      Random & 77.02 & 84.13 & 88.96 & 92.32 & 85.61\\
      MUIER (ALIGN)  & 78.78 &	84.10	& 89.22 &	92.07 & 86.04 \\
      MSIER (ALIGN) & 78.17 &	83.79	& 89.38 &	94.02 & 86.34 \\
      \bottomrule
    \end{tabular}
}
\caption{Comparison of M-ICL performance of random selection, MUIER,  and MSIER method on MS COCO dataset with ALIGN model as the backbone retriever.}
\label{tab:align}
\end{table}

% \section{More Related Work}
% \label{sec:appendix::rw}
% \textbf{In-context Learning} In-context learning was first proposed by \cite{brown2020language} in their paper introducing GPT-3. It is a significant departure from the traditional learning method based on stochastic gradient descent and does not involve any parameter updates: the model needs to be provided with a few examples of the task as part of the context for text generation.
% The size of the model and training data were thought to be key to training a model with in-context learning capabilities. More recently, there has been more research on the exact causes of in-context learning. \cite{min2022metaicl} has proposed MetaICL, a meta-training framework to elicit in-context learning capabilities in text-only language models. MetaICL conditions each example with related in-context examples during training. 
% % Chan et al. [5] have investigated the distributional properties of training data for in-context learning. 
% % Their findings have shown that certain properties encourage in-context learning in language models, and massive textual data from the web used to train large language models naturally have those properties.
% In the context of MLLMs, ICL has been extended to more modalities, leading to Multimodal ICL\cite{alayrac2022flamingo, li2023otter, chen2023sinc}. 

\section{More Evaluation Examples}

\label{sec:appendix::examples}
More evaluation examples using different retrieval methods are presented here:
\begin{table*}[ht]
  \centering
  \begin{tabular}{|p{0.3\linewidth}|p{0.3\linewidth}|p{0.3\linewidth}|}
    \hline
    \textbf{RICES} & \textbf{MUIER} & \textbf{MSIER} \\
    \hline
    A male tennis player celebrating a point.  &
     A male tennis player celebrating a point on the tennis court.  &
    A man holding a tennis racket with a ball in the air on the tennis court. \\
    \hline
    \multicolumn{3}{|p{12cm}|}{\textbf{Test example}} \\
    \multicolumn{3}{|c|}{\includegraphics[width=0.5\linewidth, height=8cm]{figs/COCO_val2014_000000361147.jpg}} \\
    % \multicolumn{4}{|p{12cm}|}{OF-9B generations: A yellow labrador retriever running with a ball.} \\
    \multicolumn{3}{|c|}{\textbf{GROUND TRUTH:} \textit{A tennis player is standing on a tennis court with a racquet in his hand.}} \\
    \hline
  \end{tabular}
\end{table*}

\begin{table*}[ht]
  \centering
  \begin{tabular}{|p{0.3\linewidth}|p{0.3\linewidth}|p{0.3\linewidth}|}
    \hline
    \textbf{RICES} & \textbf{MUIER} & \textbf{MSIER} \\
    \hline
    A man in a blue shirt and tie.   &
    A man wearing a blue shirt with a black tie.  &
    A man in a blue shirt sits at a desk in front of a computer monitor. \\
    \hline
    \multicolumn{3}{|p{12cm}|}{\textbf{Test example}} \\
    \multicolumn{3}{|c|}{\includegraphics[width=0.5\linewidth, height=8cm]{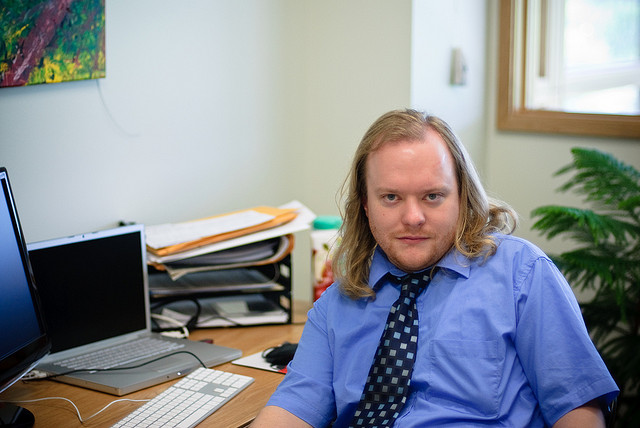}} \\
    % \multicolumn{4}{|p{12cm}|}{OF-9B generations: A yellow labrador retriever running with a ball.} \\
    \multicolumn{3}{|c|}{\textbf{GROUND TRUTH:} \textit{A man with long hair and a blue shirt sitting in front of a computer.}} \\
    \hline
  \end{tabular}
\end{table*}

\begin{table*}[ht]
  \centering
  \begin{tabular}{|p{0.3\linewidth}|p{0.3\linewidth}|p{0.3\linewidth}|}
    \hline
    \textbf{RICES} & \textbf{MUIER} & \textbf{MSIER} \\
    \hline
    A surfer standing on a surfboard.  &
     A surfer in a wetsuit holding a surfboard.   &
    A man wearing a wet suit standing on the beach holding a surfboard. \\
    \hline
    \multicolumn{3}{|p{12cm}|}{\textbf{Test example}} \\
    \multicolumn{3}{|c|}{\includegraphics[width=0.5\linewidth, height=8cm]{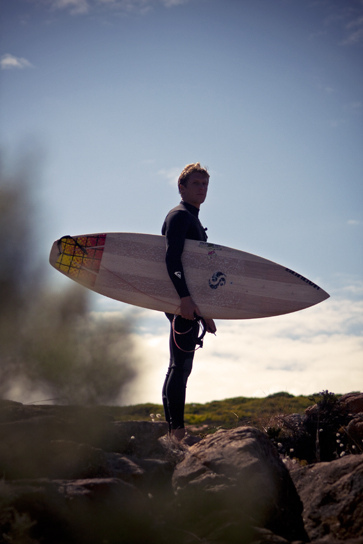}} \\
    % \multicolumn{4}{|p{12cm}|}{OF-9B generations: A yellow labrador retriever running with a ball.} \\
    \multicolumn{3}{|c|}{\textbf{GROUND TRUTH:} \textit{ A man in black wet suit holding a surfboard under his arm.}} \\
    \hline
  \end{tabular}
\end{table*}

\end{document}